\documentclass[12pt,letterpaper]{article}
\usepackage[a4paper, total={7in, 10in}]{geometry}

\usepackage{graphicx}
\usepackage{helvet}
\usepackage{authblk}
\usepackage{hyperref}
\usepackage{setspace}
\usepackage{amsmath} 
\usepackage{amssymb}
\usepackage{soul}
\usepackage{xcolor}
\DeclareRobustCommand{\hlcolor}[1]{{\sethlcolor{white}\hl{#1}}}
\DeclareRobustCommand{\hllcolor}[1]{{\sethlcolor{white}\hl{#1}}}
\DeclareRobustCommand{\hLcolor}[1]{{\sethlcolor{white}\hl{#1}}}

\usepackage[normalem]{ulem}
\usepackage[super,comma,sort&compress]  
{natbib}
\usepackage[right]{lineno} \linenumbers

\makeatletter
\renewcommand{\maketitle}{\bgroup\setlength{\parindent}{0pt}
	\begin{flushleft}
		\textbf{\@title}
		
		\@author
	\end{flushleft}\egroup}
\makeatother

\title{\Large Defect Detection in Photolithographic Patterns Using Deep Learning Models Trained on Synthetic Data}
\date{}

\author[1,*]{Prashant P. Shinde}
\author[1]{Priyadarshini P. Pai}
\author[1,**]{Shashishekar P. Adiga}
\author[1]{K. Subramanya Mayya}
\author[2]{Yongbeom Seo}
\author[2]{Myungsoo Hwang}
\author[2]{Heeyoung Go}
\author[2]{Changmin Park}

\affil[1]{NextGen Projects (SAIT-India), Samsung Semiconductor India Research, Bangalore, 560048, India}
\affil[2]{Foundry Process Development Team, Semiconductor R\&D Center, Samsung Electronics, Seoul, Korea}

\affil[*]{Correspondence: prash.shinde@samsung.com}
\affil[**]{Correspondence: shashi.adiga@samsung.com}

\begin{document}
	\doublespacing
	\nolinenumbers
		
	\maketitle
	
	\section*{Abstract}
	
	In the photolithographic process vital to semiconductor manufacturing, various types of defects appear during EUV pattering. Due to ever-shrinking pattern size, these defects are extremely small and cause false or missed detection during inspection. Specifically, the lack of defect-annotated quality data with good representation of smaller defects has prohibited deployment of deep learning based defect detection models in fabrication lines. To resolve the problem of data unavailability, we artificially generate scanning electron microscopy (SEM) images of line patterns with known distribution of defects and autonomously annotate them. We then employ state-of-the-art object detection models to investigate defect detection \hlcolor{performance} as a function of defect size, much smaller than the pitch width. We find that the real-time object detector YOLOv8 has the best \hlcolor{mean average precision of 96\% as compared to EfficientNet, 83\%, and SSD, 77\%}, with the ability to detect smaller defects. We report the smallest defect size that can be detected reliably. When tested on real SEM data, the YOLOv8 model \hlcolor{correctly detected 84.6\% of Bridge defects and 78.3\% of Break defects across all relevant instances}. These promising results suggest that synthetic data can be used as an alternative to real-world data in order to develop robust machine-learning models.
	
	\section*{Keywords}
	
	
	Scanning Electron Microscopy, Data Generation, Machine Learning, Defect Detection.
	
	\section*{Introduction}
	
	The shrinkage of pattern dimensions in advanced semiconductor nodes necessitate more complex fabrication processes, which typically have high propensity for nanoscale defects. Notwithstanding the efforts made towards detection and root-cause-analysis to improve yield\cite{R1, R4, R5}, these small-size defects are difficult to detect even with high-sensitivity microscopy inspection tools such as scanning electron microscopy, because of low-contrast compared to the background and high noise level.
	
\begin{figure}[ht]
	\begin{center}
		\includegraphics[width = 0.60\textwidth]{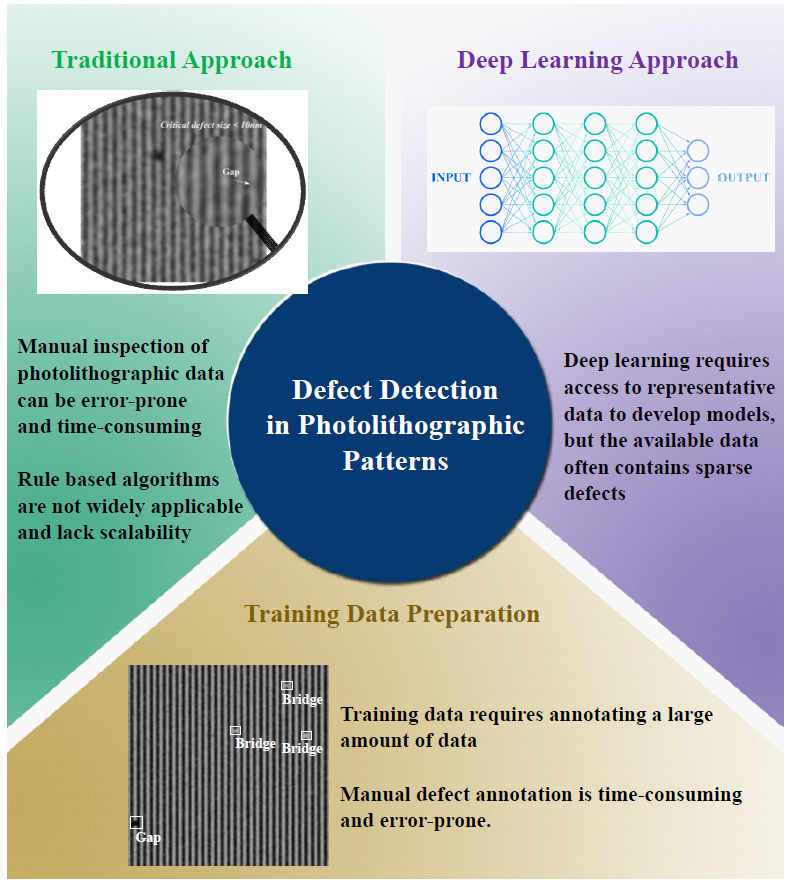}
		{\caption{Overview of defect detection challenge}\label{fig:FigGA}}
	\end{center}
	\end{figure}

	\noindent \hlcolor{Manual or rule-based defect detection is often error-prone and time-consuming}. \hlcolor{As a result, state-of-the-art machine learning and deep learning (ML/DL) techniques for automated defect detection are gaining prominence in defect metrology}\cite{R7, R8, R9}. For successful development and deployment of ML/DL defect detection models, two key challenges remain to be solved. The first challenge concerns the effectiveness of ML/DL models in detecting small-size objects. Detecting small objects is a challenging task in computer vision due to their small visual representation\cite{R10, R11} and that there is further reduction in size of the defect when passed through successive convolutional layers. This problem is magnified when the images contain a lot of noise, which is often the case in SEM images. Given the importance of detection of small defects in advanced nodes \cite{R12}, the questions that naturally arise are:
	\begin{enumerate}
		\item What is the smallest defect that can be detected reliably?
		\item What ML/DL models perform better at detecting small-size defects?
	\end{enumerate}
	These important questions need to be addressed before DL based defect detectors can be deployed in the fabrication line. The second and equally significant challenge is the availability of representative and annotated defect metrology data for training DL models. This problem primarily stems because (i) the acquisition of SEM images is slow and expensive; (ii) the occurrence of defects is rare and often type-unbalanced; collecting sufficient representative data for predictive modelling is arduous; (iii) and more importantly, manual annotation of such defects is time-consuming and not scalable to large amounts of data. With increasing complexity of defects under low contrast and high level of noise, annotating small defects is in general challenging. Figure~\ref{fig:FigGA} outlines these challenges.
	\vspace{0.5cm}
	
	\noindent To address these challenges and to answer the above questions, we consider the problem of defect detection in photolithographic line patterns that typically consist of a series of parallel lines of photoresist. Since photoresist features in advanced nodes continue to shrink, the defects in these line patterns, that include breaks in the line and bridges between two lines, are nanometer scale and require SEM imaging for detection. The small size of the defects and inherent noisiness of SEM images pose a serious challenge to reliable defect detection using ML/DL. Further, as the target defect concentration in next generation photolithography patterns is less than one defect in a trillion pattern features, imaging enough defects, for training ML/DL models, is a time consuming process and add to that the manual efforts involved in annotating these defects.
	
	\noindent In this work, we first tackle the issues of data unavailability and annotation by generating virtual/synthetic data. Simultaneous generation of images and automated annotation of defects is an alternative way for generating balanced data with the eventual goal of developing a robust performing deep learning model. We then apply state-of-the-art object detectors and compare their ability to detect defects by analyzing \hlcolor{true positive rate or hit rate} as a function of defect size. \hlcolor{In the end, we report performance of a state-of-the-art real-time defect detection model on real-world data.}

	\noindent In recent years, there has been a proliferation of deep learning methods in almost every sector of business and research. Many deep learning based studies achieving promising detection have reported results on specific datasets. \hllcolor{Evanschitzky et al}\cite{R15} \hllcolor{reported the detection of mask pattern defects and the determination of exact defect shapes from SEM images using a combination of several deep learning networks and analytical methods. Slicing Aided Hyper Inference framework improved the performance of detection models on smaller objects by dividing an image into overlapping patches, enlarging these patches, and refining the detection process}\cite{R16}. A recent study on defect detection in transmission electron microscopy images\cite{R17, R18, R19} using machine learning was focused on detection of point defects in noisy images. Their proposed method involves using a PCA reconstruction to generate a residual image and then a deep neural network to detect and classify the presence of an anomaly in the residual image. \hllcolor{A study utilized conditional generative adversarial network (cGAN) to generate artificial SEM images and balance the dataset, reporting an improvement in detection performance on a real industrial dataset} \cite{R20}. \hllcolor{The cGAN employed a key point estimation method, such as heatmaps, instead of anchor boxes for object detection. In their study, the defects were relatively large compared to the image size}. 
	 To achieve both high detection accuracy and high speed, single-stage target detection algorithms such as YOLO\cite{R21} and its variants are developed and widely applied for locating smaller targets. Moreover, YOLO variants have been reported to achieve significant performance gain in scenarios with limited data availability and highly imbalanced datasets\cite{R22, R23}. The proposed methods utilize data augmentation techniques to evenly distribute the classes for training and validation. \hllcolor{In general, training with balanced class data typically results in improved model performance} \cite{R24}. \hllcolor{Dey et al.} \cite{R22} \hllcolor{extracted various small patches from the original SEM images. Prior to the extraction of the patches, human intervention was necessary to annotate/label the acquired data. The authors then used Denoising Diffusion Probabilistic Models (DDPM) to generate realistic semiconductor wafer SEM images, thereby increasing defect inspection training data. Note that data augmentation alone appears insufficient to address the data availability issue in semiconductor defect inspection, particularly for small defects, emphasizing the need for a method to create new training data without relying on SEM data. Our proposed approach generates synthetic images without the need for SEM data. A pattern of grey and dark regions or pixels with predefined widths is repeated throughout the entire width and height of an image. Further details are provided in the following section.}
	
	\noindent Further, various inputs have also been explored, such as-for scanning transmission electron microscopy\cite{R17}, optical wafer images\cite{R26} and printed circuit boards (PCB) images captured using industrial camera\cite{R27}. A detailed review of the state-of-the-art machine vision based defect detection is available for further reference\cite{R28}. Recent efforts\cite{R29, R30} on detecting defects in wafers have shown applications of machine learning in semiconductor manufacturing. \hllcolor{An effective  ensembling strategy} \cite{R31, R32}, \hllcolor{combining predictions from different models, was shown to improve detection precision for large defects in SEM images. Their models demonstrated lower mAP when detecting challenging small defects. An ensemble of deep learning models could enhance the mAP.} A segmentation network was used to find defects in SEM images with complex structure \cite{R33}. \hlcolor{Research} \cite{R34, R35} \hlcolor{on object detection has often emphasized detecting objects with larger aspect ratios within images, while smaller objects tend to be more challenging due to their finer details and lower resolution within specific datasets.} Because small objects occupy small area of an image, they yield weak features to support localization. Based on the available important features, the potential of machine learning techniques is being explored to find critical small-size defects in semiconductor data. However, collecting enough semiconductor data cannot guarantee detection of stochastic defects due to their unpredictable and random nature. Stochastic defects can happen irrespective of pattern type.
	
	\section*{Method details}

\label{sec:methods}

Good quality real SEM data with annotations is essential for developing high performing deep learning models, which are applicable to specific tasks with specific requirements. \hlcolor{Generative adversarial networks (GANs)} \cite{R36, R22} and style transfer \cite{R37, R38} methods can generate a new set of high-quality images with multiple styles from training examples. However, these methods require large amount of diverse data to produce good results. Style transfer can produce images that look unnatural and artificial. This is because the style of a source image is transferred to the target image without understanding the content of the images which can lead to strange results for SEM images where close attention to details is crucial. In addition, these methods are incapable of annotating objects present in the output images. The generated images need to be manually annotated which is laborious. It is time-consuming to collect high-quality SEM data from semiconductor fabrication lines, and can be expensive.

\noindent Synthetic data, in contrast, with accurate annotations can be created manually or generated automatically for variety of tasks. As an example, we generate good quality SEM images of lithographic line-space patterns with known defects annotated for bounding box. A space in a photoresist line forms a ‘Break’ defect and a connection between adjacent photoresist lines forms a ‘Bridge’ defect. We simulate Break and Bridge defects of various sizes normalized by the width of lithographic line. Note here that the pitch-to-line-width ratio is close to two. \hlcolor{Smaller bridge and break defects occur rarely in real-world data, making it challenging to collect enough representative data for them.} \hLcolor{We found only Break and Bridge defects in our actual SEM data.} \hlcolor{Therefore, we focus on modelling these defects. These smaller defects are also more difficult to annotate compared to larger defects that are easier to identify.}

\begin{figure}[h!]
	\begin{center}
		\includegraphics[width = 0.67\textwidth]{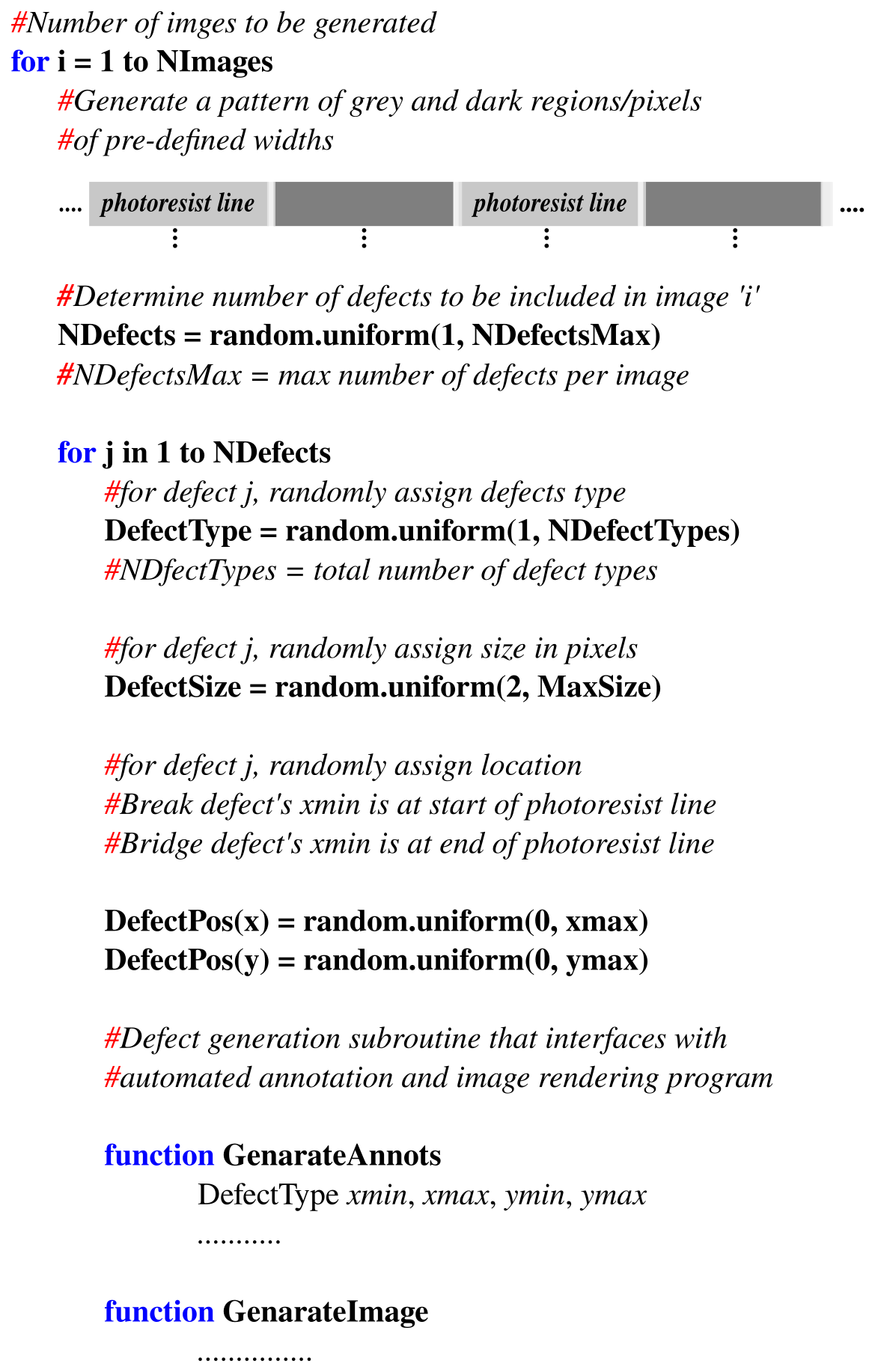}
		{\caption{Pseudocode to generate synthetic SEM images with various defects distributed uniformly.}\label{fig:FigA2}}
	\end{center}
\end{figure}

\begin{figure}[ht]
	\begin{center}
		\includegraphics[width = 0.67\textwidth]{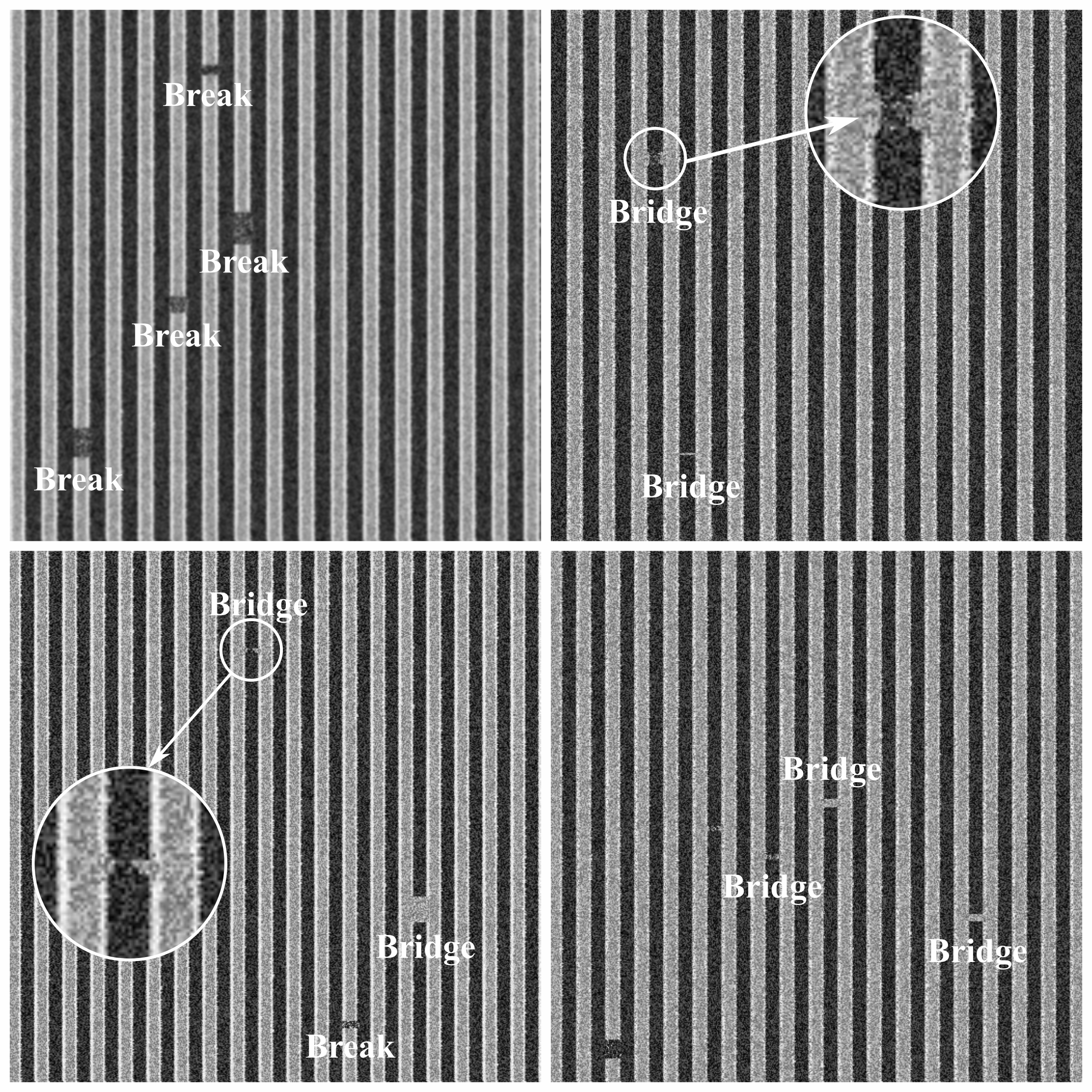}
		{\caption{Examples of synthetic SEM images of lithographic line-space patterns with Break and Bridge defects. Enlarged regions indicate probable Bridge defects. Defects of various types and sizes are shown.}\label{fig:FigA3}}
	\end{center}
\end{figure}

\noindent Figure~\ref{fig:FigA2} shows the pseudocode to generate class balanced SEM data. Our framework to generate SEM images facilitates user input, including the number of images to be generated (NImages), image resolution in pixels, and maximum number of Break and/or Bridge defects (NDefectsMax) to be introduced in an image. For an image, we first generate a pattern of alternate grey and dark regions of pre-defined widths in pixels over entire column (width) of an image. A bright boundary on either sides of these regions forms an edge of photoresist line. The intensity of a pixel in grey regions is determined using a uniform random number within a given range between a minimum and a maximum value. 
The pattern of grey and dark regions repeats over entire rows (height) of an image. Figure~\ref{fig:FigA3} shows examples of the generated synthetic SEM images of lithographic line-space patterns. By varying the widths of grey and dark regions, one can control the pitch width. A Bridge defect is a group of unwanted grey pixels in a dark region whereas a Break defect is a group of unwanted dark pixels in a grey region. A Bridge defect acts as a connection between adjacent lines. The location of a defect is pre-determined using a uniform random number generator.

\noindent Each image has at least one defect with distinct type (DefectType), \hllcolor{size (DefectSize; in pixels along the y-axis)}\hlcolor{, and location (DefectPos) determined using uniform random numbers}. The maximum number of defects per image is set to five. Use of uniform random numbers ensures uniform distribution of multiple defects of different sizes. We write a defect profile (size and location) in a text file containing information about defect type and bounding box coordinates (\textit{xmin}, \textit{ymin}, \textit{xmax}, \textit{ymax}) that outline the defect in the box. These coordinates contain two pairs of whole numbers. The first pair (\textit{xmin}, \textit{ymin}) indicates coordinates of the top left corner of the box and the second pair (\textit{xmax}, \textit{ymax}) indicates the bottom right corner of the box. All synthetic SEM images are autonomously annotated for bounding box. The annotations, defect type and bounding box, can be stored in standard formats (Pascal VOC, YOLO, etc.) to be used for training object detection models. The accurately annotated data must be supplied as input in the form that an ML algorithm understands. The autonomous annotation of SEM data creates a highly accurate ground truth directly influencing the performance of ML models. There are a few benefits of our strategy of automated annotations and image generations as follows:
\begin{enumerate}
	\item Automated annotations strategy can be tuned to specific process conditions/variations specific to a fab.
	\item It offers a very efficient method to generate automatically annotated class balanced data – reducing cost and increasing defect detection efficiency.
	\item ML/DL models developed with the synthetic data are useful for detecting defects in real SEM or optical metrology data.
	\item The strategy of generating data, developing ML/DL models, and applying it to real data can be integrated into semiconductor metrology pipeline for defect root-cause analysis and yield improvement. The proposed solution will reduce thousands of man-hours.	
\end{enumerate}

\begin{table}[h]
	\begin{center}
		\begin{tabular}{|*{3}{c|} }
			\hline
			Dataset    & \multicolumn{2}{c|}{Defect Type}\\
			& \multicolumn{2}{c|}{Break\hspace{57pt}Bridge}\\
			\hline
			\hspace{17pt} Training \hspace{17pt}  & \hspace{17pt}  10425  \hspace{17pt} & \hspace{17pt}  10390 \hspace{17pt} \\
			\hline	
			Validation   &   2990  &   3023  \\
			\hline
			Test   &   1502  &   1506  \\
			\hline
		\end{tabular}
		\caption{Distribution of defects in various datasets.}
		\label{tab1}
	\end{center}
\end{table}

\noindent Table~\ref{tab1} indicates a uniform distribution of Break and Bridge defects in the generated data. The generated dataset comprises 10000 synthetic SEM images of lithographic line-space patterns. All these images are of size 512x512. The number of examples per class is $\sim$50\%, which is expected due to equal likelihood of happening. Out of 10000 images, 70\% (7000) images were used for training, 20\% (2000) images for validation, and the remaining 10\% (1000) images for testing. Nearly 1050 defects of every size are present in the training dataset. The number of images or number of examples per class considered in the present work are sufficient for developing effective defect detection models. Figure~\ref{fig:FigA3} shows examples of generated images with Bridge and Break defects of various sizes. The size of Break and Bridge defects in the generated data varies from 0.1 to 2 times the width of photoresist line. Enlarged regions shown in these example images indicate probable Bridge defects.

\noindent Microscopy images are prone to various sources of noise, inherent to visualization/inspection techniques. Noise in real microscopy images is highly complex and intensity dependent. The exact noise model is unknown and interests a lot of research work \cite{Noise0, Noise1}. In the present work, we use uniform random numbers to give random variation to brightness/darkness of the grey and dark regions in SEM images. The random noise can be characterized by its unpredictability. In this respect, we match the signal-to-noise ratio of the generated microscopy images with the available in-house real microscopy images. The coefficient of variation based signal-to-noise ratio for synthetic SEM images varies from 3.7 to 9.0 dB \hllcolor{(4.1 to 8 dB for real images)}. Introduction of noise adds variability to images resulting in preventing overgeneralization of ML/DL models. \hllcolor{The brightness, RMS contrast, line width, and pitch width of synthetic images vary within the ranges [96-153], [33-77], [10-23], and [19-47], respectively, compared to [102-141], [38-54], [11-25], and [21-52] for real images.} \hLcolor{Our proposed approach generates synthetic images without relying on real SEM data, making it fundamentally different from conventional methods that require real SEM inputs. As a result, a direct comparison between synthetically generated images and real SEM images is not feasible, as there is no corresponding ground truth for a one-to-one 	evaluation.} The performance of ML/DL models is evaluated based on the following metrics.

\subsection*{Performance evaluation metrics}
In the realm of object detection, the most commonly used metrics to assess ML model’s performance are precision, recall, mean average precision, accuracy, and F-score. Precision is the ratio of True Positives to all Positives. It is a measure of quality and is calculated as,
\begin{equation}
	Precision=\frac{TP}{TP+FP}
\end{equation}
Where, TP and FP stand for True Positives and False Positives, respectively. The recall is a measure of quantity. It is calculated as follows.
\begin{equation}
	Recall \; (TPR) = \frac{TP}{TP+FN}
\end{equation}
Where, FN stands for False Negatives. Recall indicates how accurately our model is able to identify the relevant data. It is also called as Sensitivity \hlcolor{ or hit rate or true positive rate (TPR)}. Another simplest metrics of all is Accuracy. It is the ratio of the total number of correct predictions to the total number predictions.
\begin{equation}
	\begin{aligned}
		Accuracy &= \frac{TP+TN}{TP+FP+TN+FN} \cdot 100 = \frac{Correct\ Predictions}{Total\ Records} \cdot 100\\
	\end{aligned}
\end{equation}
\hlcolor{Where, TN is true negative. In the context of object detection, it is important to note that a true negative is not applicable as there are an infinite number of bounding boxes that should not be detected in any given image. One should avoid using any metric that relies on TN}\cite{NoTN}. \hlcolor{In the later discussions, we use TPR to analyze how well a model identifies true positives.} F-score or F1-score is also an important metric to measure the performance of a model. It is the weighted average (or harmonic mean) of precision and recall.
\begin{equation}
	F-score=2 \cdot \frac{Precision \cdot Recall}{Precision + Recall}
\end{equation}
It gives a better measure since it penalizes the extreme values. 
	\section*{Results and discussion}
	\subsection*{Model training on synthetic data}
	In this section, we explore the scope of training the state-of-the-art (SOTA) deep learning models with synthetic data for localizing defects and identifying their types. Our primary objective is to demonstrate that synthetic models can really bring value to this field and address the critical issue of unavailability of annotated data. We focus on predicting a bounding box for the region-of-interest in virtual SEM images. We train the state-of-the-art object detection models namely, YOLOv8 (large)\cite{YOLOv8}, SSD\cite{SSD}, and EfficientNet\cite{EffNet} with the synthetic data that consists of two types of defects. These object detection models are useful for real-time applications with fast inference. To train these models, a desktop with \hlcolor{a} single NVIDIA GeForce RTX 3090 with 24 GB graphics card is used.
	
	\noindent \hlcolor{The initial} values of parameters in deep neural networks are defined using \hlcolor{the Kaiming} weight initialization technique, which is crucial for training models on a dataset. We tune adjustable parameters to develop a model with optimal performance. To train the models, a learning rate of 2.1E-4 along with momentum of 0.913 and weight decay of 4.9E-5 are used. We use a batch size of 24 while training and validating a model. To judge objectively, whether the model predicted a bounding box location correctly or not, we define an intersection over union (IoU) threshold ($\geq$ 0.5). The threshold determines if the object detected is valid or not. It represents an overlap between the predicted box and the ground truth, and is determined positive if an IoU score is greater than or equal to the threshold.
	A different threshold can be chosen depending on specific priorities.
	
\begin{table}[h!]
	\begin{center}
		\begin{tabular}{|*{7}{c|}}
			\hline
			\multicolumn{1}{|c|}{Metric} & \multicolumn{2}{c|}{YOLOv8}
			& \multicolumn{2}{c|}{EfficientNet}
			& \multicolumn{2}{c|}{SSD}        \\
			& \multicolumn{2}{c|}{Bridge\hspace{23pt}Break} & \multicolumn{2}{c|}{Bridge\hspace{23pt}Break} & \multicolumn{2}{c|}{Bridge\hspace{23pt}Break}\\
			\hline
			TP & 1455 \hspace*{11pt} & 1379 & 1250 \hspace*{11pt} & 1137 & 1061 \hspace*{11pt} & 911 \\
			\hline
			FP & 0 \hspace*{11pt} & 107 & 177 \hspace*{11pt} & 297 & 259 \hspace*{11pt} & 309 \\
			\hline
			FN & 51 \hspace*{11pt} & 123 & 256 \hspace*{11pt} & 365 & 445 \hspace*{11pt} & 591 \\
			\hline
			mAP        & \multicolumn{2}{c|} {0.96}         &      \multicolumn{2}{c|}{0.83}&       \multicolumn{2}{c|}{0.77}   \\
			\hline
			F-score (mean)       & \multicolumn{2}{c|} {0.95}         &      \multicolumn{2}{c|}{0.81}&    \multicolumn{2}{c|}{0.71}   \\
			\hline
			Inference time (ms)        & \multicolumn{2}{c|} {7.3}         &      \multicolumn{2}{c|}{21}&       \multicolumn{2}{c|}{23}   \\
			\hline
		\end{tabular}
		\caption{\hlcolor{Mean average precision (mAP), F-score, and Inference time for defect detection using SOTA models trained and tested on synthetic data.}}
		\label{tab2}
	\end{center}
\end{table}
	
	\noindent We develop models with these set of parameters and estimate important metrics for evaluating the performance of models considered in this work. Table~\ref{tab2} indicates \hlcolor{important metrics} predicted by the models on the \hlcolor{test} dataset. \hlcolor{FP and FN occur when the outcome of an experiment does not accurately reflect what happened in reality. TP indicates correct predictions by a model. The concept of true negative (TN) in object detection can be ambiguous, as object detection typically focuses on evaluating TP, FP, and FN, with mAP being the standard metric for assessing detection accuracy. TNs are generally not used because they do not clearly map onto this task} \cite{NoTN}\hlcolor{, unlike in binary classification where they indicate absence of a predicted object in an irrelevant region.}
	
	\noindent The mAP is a popular metric used to evaluate the robustness of object detection models. \hlcolor{It is generally used because it provides a more meaningful assessment of performance in object detection by evaluating both localization and classification. In this study, precision and recall are computed for each defect type separately and the mAP is calculated as the mean of average precision for each type.} The F-score conveys the balance between precision and recall when there is asymmetry in the data. The higher the F-score, the better is the performance of the model. Note here that the choice of an evaluation metric depends on the type of problem\cite{NoTN, R45}. The mAP and F-score are higher for YOLOv8 model when compared to SSD and EfficientNet models. The mAP (F-score) using YOLOv8 model is 0.96 (0.95). We find no significant change in the metrics upon reducing the size of training dataset to 1000 images. This is due to the fact that two training datasets of different sizes have same underlying distribution. SSD model has reached the mAP \hlcolor{(F-score) of 0.77 (0.71)}. This could be because SSD incurs degraded performance for defect detection in low-quality noisy images. A recent study \cite{R46} reported poor performance of SSD model on noisy images. In addition, the mAP reported using the SSD model was very low. In the present work, EfficientNet model predicted the mAP and F-score of 0.83 and 0.81, respectively. These metrics values are higher than SSD values, but are lower than the values predicted by YOLOv8. The better performance of YOLOv8 indicates that the model is more robust against image noise. In the present study, we obtained a mAP of 0.96 using YOLOv8 model trained with synthetic data. This higher mAP value is due to low number of false positives. It is important to note here that there exists no means to perform fair comparative analysis between different published works unless standard models and datasets are made available for testing.

\begin{figure*}[h!]
	\begin{center}
		\includegraphics[width=0.75\textwidth]{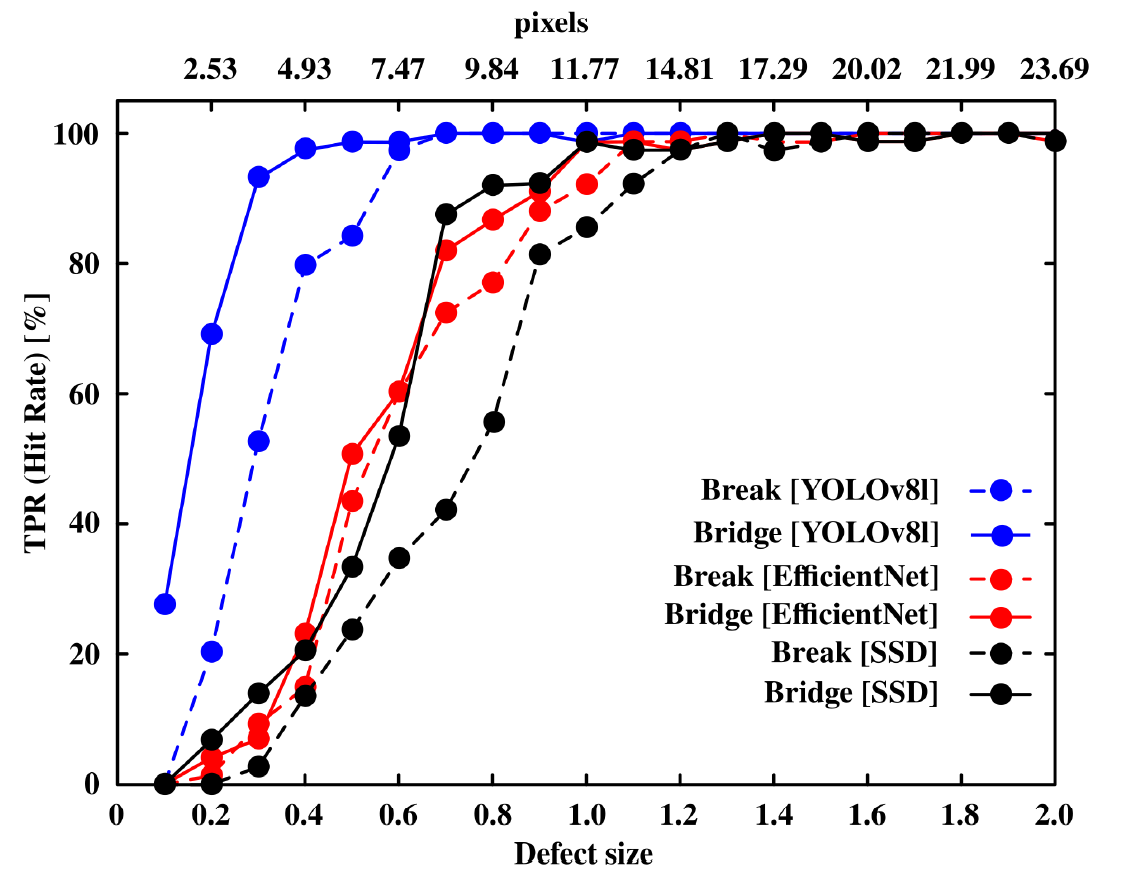}
		{\caption{\hlcolor{True positive rate (Hit Rate)} of YOLOv8, SSD, and EfficientNet models for Break and Bridge defects. Defect size is normalized by half-pitch width of the line pattern.}\label{fig:FigA4}}
	\end{center}
\end{figure*}
	
	\noindent \hlcolor{Missing a defect in semiconductor manufacturing can be highly costly or unacceptable. While the ultimate goal is zero defects, some level of defect escape might be tolerated depending on the nature of the defect, the criticality of the product, and customer requirements. However, continuous improvement efforts focus on minimizing defect escapes to ensure high product quality and reliability. In this respect, we analyze how well a model identifies correct positives.} To get insights into how effective these models \hlcolor{are} in detecting small defects, we analyzed the \hlcolor{TPR or Hit Rate} as a function of defect size normalized by the width of line pattern. Figure~\ref{fig:FigA4} shows the \hlcolor{TPR (Hit Rate)} as a function of defect size using YOLOv8, SSD, and EfficientNet models. YOLOv8 predicts an overall \hlcolor{TPR} of 96.6\% and 91.8\% for Bridge and Break defects, respectively. The average \hlcolor{hit rate} for both defects of all sizes is 94.2\%. Its \hlcolor{TPR} for both Bridge and Break defects of sizes greater than 0.50 times the width of line patterns is close to 99\%. Similarly, EfficientNet and SSD models show high \hlcolor{TPR} of 93.2\% and 89.9\%, respectively, when the size of defects is greater than 0.50 times the width of line pattern. As the size of defects decreases below 0.50, the \hlcolor{TPR} using YOLOv8 drops significantly (see Blue curves in Figure~\ref{fig:FigA4}). The \hlcolor{hit rate} predicted by \hlcolor{the} YOLOv8 model for Bridge and Break defects of sizes less than equal to 0.5 is close to 68.2\%.  YOLOv8 reaches its best \hlcolor{TPR} for size close to 0.5 \hlcolor{($\sim$7 pixels)} whereas SSD and EfficientNet models reach maximum \hlcolor{TPR} for size close to 1.2 \hlcolor{($\sim$15 pixels)}. For defects of size less than 0.5 times the width of line patterns, the performance of EfficientNet and SSD models is very poor. These models show \hlcolor{hit rate} less than 15\% for smaller defects. \hlcolor{As mentioned previously}, lower mAP and F-score metrics indicate poor performance of EfficientNet and SSD models compared to YOLOv8.
	
	\begin{figure}[h!]
		\begin{center}
			\includegraphics[width = 0.40\textwidth]{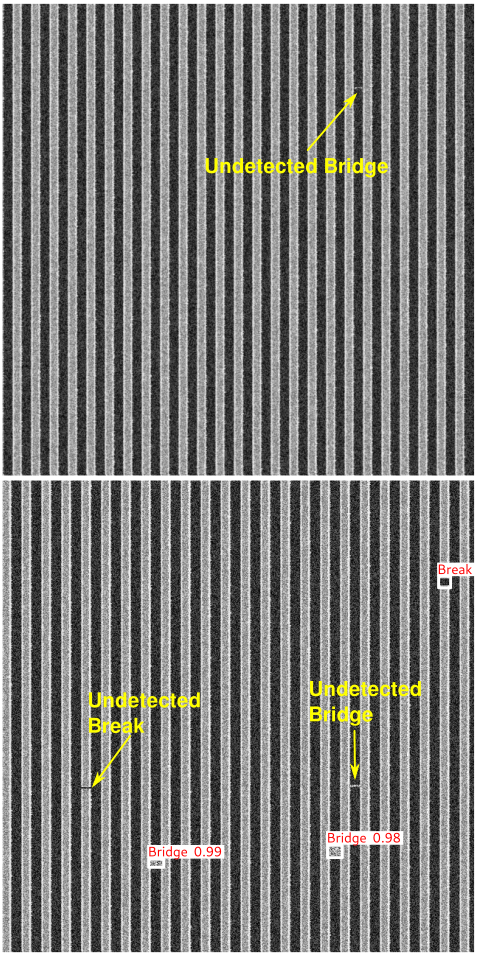}
			{\caption{Defect detection using YOLOv8 model trained on synthetic SEM images of lithographic line patterns with Break and Bridge defects. Detected defects are marked with boxes and arrows indicate undetected defects.}\label{fig:FigA5}}
		\end{center}
	\end{figure}
	
	\noindent The \hlcolor{true positive rate} using the models to detect small-size defects \hlcolor{(size $\sim$0.20 $\approx$ 3 pixels)} is very low; meaning those defects are hard to detect. For small-size defects, the resulting features are \hlcolor{weak and} inadequate for efficient learning and detection by a \hlcolor{neural network. Moreover, due to the presence of noise, the trained models greatly lose important information on the significant features of small size defects.} In turn, the models miss the small-size defects. Our results indicate that the detectable size of defects in synthetic SEM images with realistic noise is close to 0.50 times \hlcolor{($<$10 pixels)} the width of line patterns. YOLOv8 easily detects both Break and Bridge defects of moderate to larger size. However, it is less sensitive to small targets that yield weak features. Further improvement is possible in detecting small-size defects when the models are trained with appropriately modeled noise.
	
	\noindent In addition, the inference time — the amount of time taken by a model to process unseen data and make predictions — is another important metric to compare the performance of object detection models. It is important to have low inference time for an efficacious model. Our YOLOv8 model spent 7.3 ms, on average, to detect defects in an image. \hlcolor{YOLOv8 prioritizes speed, making it well-suited for real-time applications like semiconductor manufacturing.} EfficientNet and SSD models were three times slower in detecting defects (see Table~\ref{tab2}). Both these models took similar time per image.
	
	\noindent Figure~\ref{fig:FigA5} shows example images after defect detection using the YOLOv8 model. A rectangular bounding box around each defect of interest indicates the size and position of the defect in the image. Each predicted bounding box provides vital information on the location of defects, which is as equally important as detecting types of defects. Such crucial information can help in understanding the root-cause analysis of defects and the process steps. The model missed a few small-size defects of Break and Bridge types. Those defects are indicated as `Undetected'. For such undetected defects, the resulting features are inadequate for efficient learning.
	
	\subsection*{Model testing on real-world data}
\begin{center}
		\begin{figure*}[h!]
			\includegraphics[width=\textwidth]{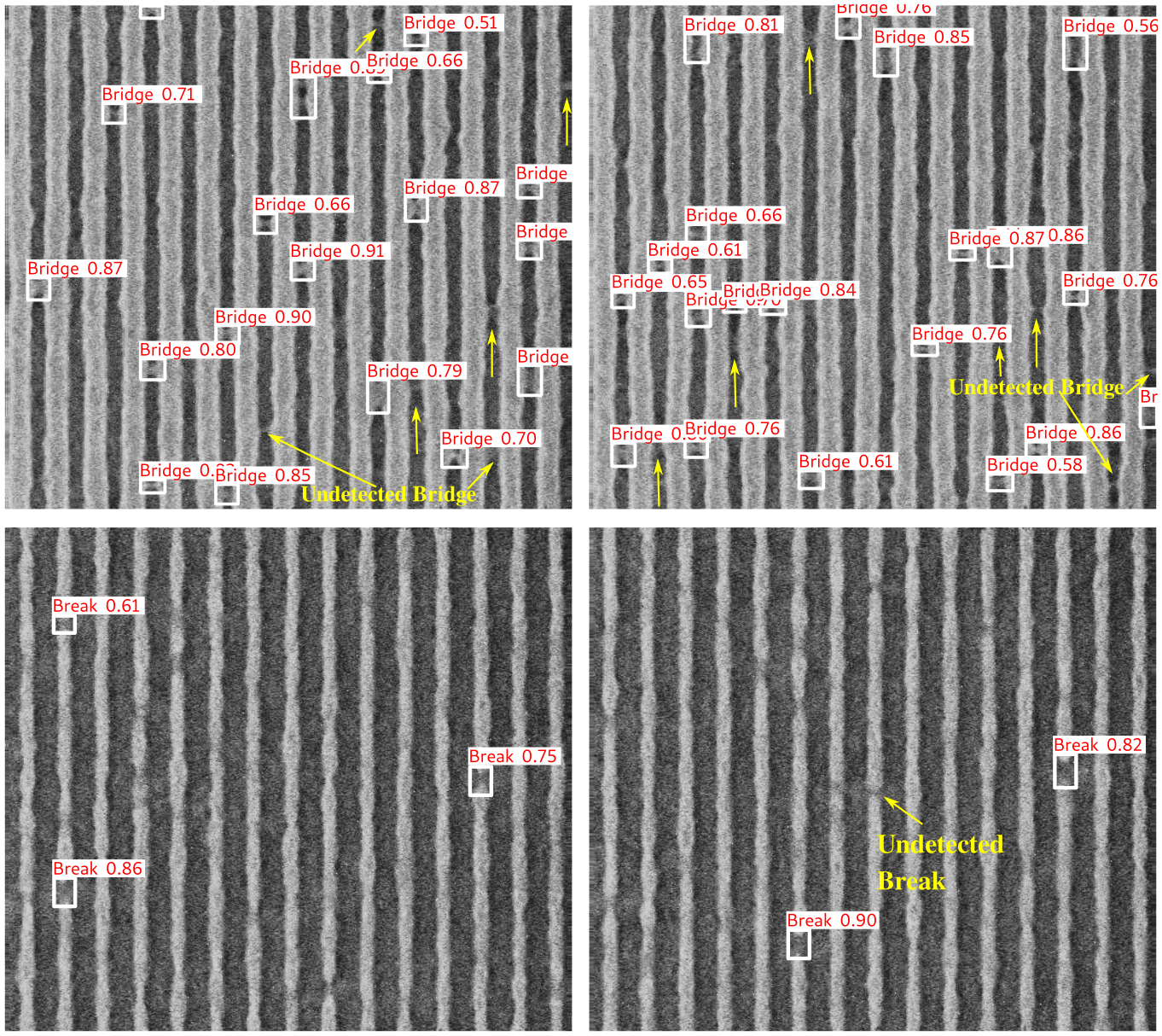}
			\caption{Defect detection using synthetic-data-trained YOLOv8 model on real-world data.  
				The rectangular boxes show examples of predicted bounding boxes and arrows indicate a few examples of undetected defects.}\label{fig:FigA6}
		\end{figure*}
	\end{center}

	By testing synthetic-data-trained YOLOv8 model on real-world data, we can effectively assess the quality of the synthetic data in terms of how closely it represents the real-world data. The performance of our best model, YOLOv8, can tell us how well the synthetic data mimics the patterns present in real data. Figure~\ref{fig:FigA6} shows defect detection on real-world data using YOLOv8 model developed on synthetic data. A rectangular bounding box indicates size and location of a defect detected by YOLOv8. Our model correctly identified regions-of-interests (ROI) with similar brightness and contrast as that of photoresist line. Object's brightness is one of the most important factors that facilitates its detection. Object detection/recognition algorithms identify objects by scrutinizing their visual characteristics. For large objects, certain features might dominate over other crucial aspects. This may affect the performance of object detector even for large size defects. \hlcolor{We see in Figure}~\ref{fig:FigA6} \hlcolor{that some  defects (indicated by yellow arrows) with similar size, brightness, and shape went undetected. The performance of YOLOv8 model varied based on factors like defect size, brightness, contrast, and shape complexity.}
	
	\noindent In deep neural networks, as the images are passed through different layers, the limited pixel representation of small objects gets reduced in spatial resolution. Since the strategy is to extract only high-level features from the input data, it may inadvertently cause small objects to be missed from the network's detection capabilities. Furthermore, dimensionality reduction in a network may cause elimination of critical details of small objects. In the end, object detection models progressively reduce spatial resolution to form high-level features for detection. \hlcolor{Such inadequate information complicates recognition of small defects. Additionally, tiny defects often have reduced visibility, causing important features to become imperceptible.}
	
	\noindent Our synthetic-data-trained YOLOv8 model \hlcolor{correctly found 84.6\% Bridge defects in real data. For Break defects, the correct detections are at 78.3\% across all relevant instances} (see Table~\ref{tab3}). \hlcolor{Our model demonstrates an average hit rate of 81.2\% for both defects.} It is hard to distinguish between `line-thinning' and `Break' defects, especially in the presence of noise. Presence of high noise level affects detection of small/tiny defects. \hlcolor{The performance of a model} can further be improved by simulating different lighting conditions by adjusting local brightness and contrast levels of ROI. In addition, a model trained with synthetic data could provide better detection \hlcolor{performance} if the test images provided have high resolution and have high signal-to-noise ratio. Image noise significantly diminishes important features of targets, which leads to difficulties in detection of small-size defects.
	
	\noindent In addition to \hlcolor{true positive rate}, YOLOv8 performs well with respect to inference time as well. \hlcolor{Our model demonstrated a detection time of 8.5 ms for defects in real-world data. The YOLOv8 model is indeed known for its impressive performance in terms of both inference time and accuracy for detecting defects in real-world applications.}
	
	\begin{table}[h!]
		\begin{center}
			\begin{tabular}{|*{4}{c|} }
				\hline
				Defect Type    & AP & F-score & \hlcolor{TPR or Hit Rate} (\%) \\
				\hline
				Bridge & 0.983 & 0.909 & 84.6\\
				\hline	
				Break   &   0.947  &   0.857  & 78.3\\
				\hline
			\end{tabular}
			\caption{Average precision (AP), F-score, and \hlcolor{true positive rate} for defect detection using YOLOv8 model trained on synthetic data and tested on real-world data.}
			\label{tab3}
		\end{center}
	\end{table}
	
	\noindent The use of synthetic data can solve the problem of data scarcity. Synthetically trained models can be as good as models trained on real data\cite{R47}. It also protects data privacy, making it easier to share with collaborators for further analysis and building robust models\cite{R48}. Furthermore, machine learning based automated detection and localization of defects can help in improving the yield and reducing the human intervention significantly.
	
	\section*{Conclusions}
	\hlcolor{The lack of high-quality, well-annotated real-world scanning electron microscopy data hinders the use of deep learning in semiconductor manufacturing, particularly for detecting small-size defects in advanced technology nodes. To address this, we generated autonomously annotated synthetic SEM data, which provides high-quality, diverse, and uniform data that can overcome privacy and data protection concerns. Our study demonstrates that synthetic data can significantly improve defect detection by providing reliable and comprehensive training datasets.}
	
	\noindent \hlcolor{We applied this approach to detect Break and Bridge defects in photolithographic line patterns, using state-of-the-art object detection models.} \hLcolor{The study is limited to Bridge and Break defects observed in our actual SEM data.} \hlcolor{Our results showed that YOLOv8 outperformed SSD and EfficientNet models, achieving the best mean average precision of 96\% with synthetic data. With real data, YOLOv8 correctly detected Bridge defects at 84.6\% and Break defects at 78.3\% across all relevant instances. Although detecting small defects remains challenging due to image noise and feature loss, our model successfully localized and classified defects.}
	
	\noindent \hlcolor{This work highlights the potential of synthetic data and deep learning to improve defect detection efficiency in semiconductor manufacturing, offering a cost-effective solution for generating and annotating training data.} \hLcolor{A fair comparative analysis of our synthetically-trained model can be performed if models and datasets are made available for testing. Our approach highlights the need for a method capable of generating training data independently of actual SEM data inputs. Additional studies with a broader dataset are required to confirm the presence of other potential defects.} \hlcolor{Future research could explore additional defect types and extend the dataset for semi-supervised or unsupervised defect detection applications.}
	
	
	
	\section*{Data and code availability}
	
	\hlcolor{In this work, we used Python programming language to generate synthetic SEM data. The generated synthetic data can be made publicly available in order to achieve maximum transparency. The data will be shared by the corresponding authors upon request. Experimental SEM data used for a case study cannot be released because it is confidential in nature and is property of the Company, where the authors are employees.}
	
	\section*{Acknowledgments}
	
	
	Financial support from SAIT-Korea is gratefully acknowledged.

	\section*{Declaration of interests}


	The authors declare no competing interests.
	
	\section*{Author contributions}
	
	
	P.S. and S.P.A. conceived the idea and designed the modeling work. P.S. wrote codes, performed the simulations, analyzed the data, and made the figures. P.S. and S.P.A. wrote the paper. Y.B., M. H., H. G., and C. P. captured the  scanning electron microscopy images. All authors discussed the results and commented on the manuscript at all stages.
	

	
\end{document}